\documentclass[10pt,twocolumn,letterpaper]{article}

\usepackage{wacv}              

\usepackage{algorithm}
\usepackage[noend]{algpseudocode}
\usepackage{amsfonts}
\usepackage{amsmath}
\usepackage{amssymb}
\usepackage{booktabs}
\usepackage{caption}
\usepackage{makecell}
\usepackage{multirow}
\usepackage{wrapfig}
\usepackage{graphicx}
\usepackage{subcaption}
\usepackage{xcolor}
\usepackage{pifont}

\usepackage[pagebackref,breaklinks,colorlinks]{hyperref}

\usepackage[capitalize]{cleveref}
\crefname{section}{Sec.}{Secs.}
\Crefname{section}{Section}{Sections}
\Crefname{table}{Table}{Tables}
\crefname{table}{Tab.}{Tabs.}
\usepackage[numbers]{natbib}
\usepackage[accsupp]{axessibility}

\def\wacvPaperID{15} 
\def\confName{WACV}
\def\confYear{2025}

\begin{document}

\title{OpenEMMA: Open-Source Multimodal Model for \\
End-to-End Autonomous Driving
}


\author{
Shuo Xing$^1$,
Chengyuan Qian$^1$,
Yuping Wang$^2$,
Hongyuan Hua$^3$,
Kexin Tian$^1$, \\
Yang Zhou$^1$,
Zhengzhong Tu$^1$ 
\\
\\
$^1$Texas A\&M University  \quad
$^2$University of Michigan
\quad
$^3$University of Toronto
\\
\texttt{\{shuoxing,tzz\}@tamu.edu}
}
\maketitle

\begin{abstract}
   Since the advent of Multimodal Large Language Models (MLLMs), they have made a significant impact across a wide range of real-world applications, particularly in Autonomous Driving (AD). Their ability to process complex visual data and reason about intricate driving scenarios has paved the way for a new paradigm in end-to-end AD systems. However, the progress of developing end-to-end models for AD has been slow, as existing fine-tuning methods demand substantial resources, including extensive computational power, large-scale datasets, and significant funding. Drawing inspiration from recent advancements in inference computing, we propose \textbf{OpenEMMA}, an open-source end-to-end framework based on MLLMs. By incorporating the Chain-of-Thought reasoning process, OpenEMMA achieves significant improvements compared to the baseline when leveraging a diverse range of MLLMs. Furthermore, OpenEMMA demonstrates effectiveness, generalizability, and robustness across a variety of challenging driving scenarios, offering a more efficient and effective approach to autonomous driving. We release all the codes in \url{https://github.com/taco-group/OpenEMMA}.
\end{abstract}

\section{Introduction}
\label{sec:intro}


Autonomous Driving (AD) technology has evolved rapidly in recent years, driven by advancements in artificial intelligence, sensor technology, and high-performance computing~\citep{hwang2024emma, shao2024lmdrive, wang2023eqdrive, sima2023drivelm,llvmad2025position, li2024comamba,li2024light}. However, real-world scenarios featuring unpredictable road users, dynamic traffic patterns, and diverse environmental conditions present significant challenges~\cite{xing2024autotrust}. Addressing these complexities requires sophisticated reasoning capabilities, allowing AD system to comprehend contextual information, anticipate user intentions, and make accurate real-time decisions. Traditionally, AD architectures have adopted a modular approach, with specialized components handling distinct aspects such as perception \citep{Yurtsever_2020, li2022hdmapnetonlinehdmap, lang2019pointpillarsfastencodersobject, sun2020scalabilityperceptionautonomousdriving, hwang2022cramnetcameraradarfusionrayconstrained}, mapping \citep{li2022hdmapnetonlinehdmap, tancik2022blocknerfscalablelargescene}, prediction \citep{nayakanti2022wayformermotionforecastingsimple, shi2024mtrmultiagentmotionprediction}, and planning \citep{Teng_2023}. However, while this compartmentalization aids in debugging and optimizing individual modules, it often leads to scalability issues due to inter-module communication errors and rigid, predefined interfaces that struggle to adapt to new or unforeseen conditions \citep{bansal2018chauffeurnetlearningdriveimitating, jiang2023vadvectorizedscenerepresentation, nayakanti2022wayformermotionforecastingsimple, seff2023motionlmmultiagentmotionforecasting}.

Recent advancements have seen the development of end-to-end systems that learn driving actions directly from sensor inputs, bypassing the need for symbolic interfaces and allowing for holistic optimization \citep{hu2023planningorientedautonomousdriving, zhai2023rethinkingopenloopevaluationendtoend, li2024egostatusneedopenloop}. However, these systems, often being highly specialized and trained on narrow datasets, struggle to generalize effectively across diverse and complex real-world scenarios. This is where Multimodal Large Language Models (MLLMs) come into play, offering a transformative approach with their extensive training on wide-ranging datasets that encapsulate comprehensive world knowledge and advanced reasoning abilities through mechanisms like chain-of-thought reasoning \citep{wei2023chainofthoughtpromptingelicitsreasoning, pan2023plum, chu2023survey}. Waymo's proprietary EMMA model \citep{hwang2024emma}, built upon Google's Gemini, exemplifies this trend, demonstrating significant advancements in integrating perception, decision-making, and navigation. Nevertheless, EMMA's closed nature restricts access and experimentation for the wider research community.

\begin{figure*}[htbp]
    \centering
    \includegraphics[scale=0.35]{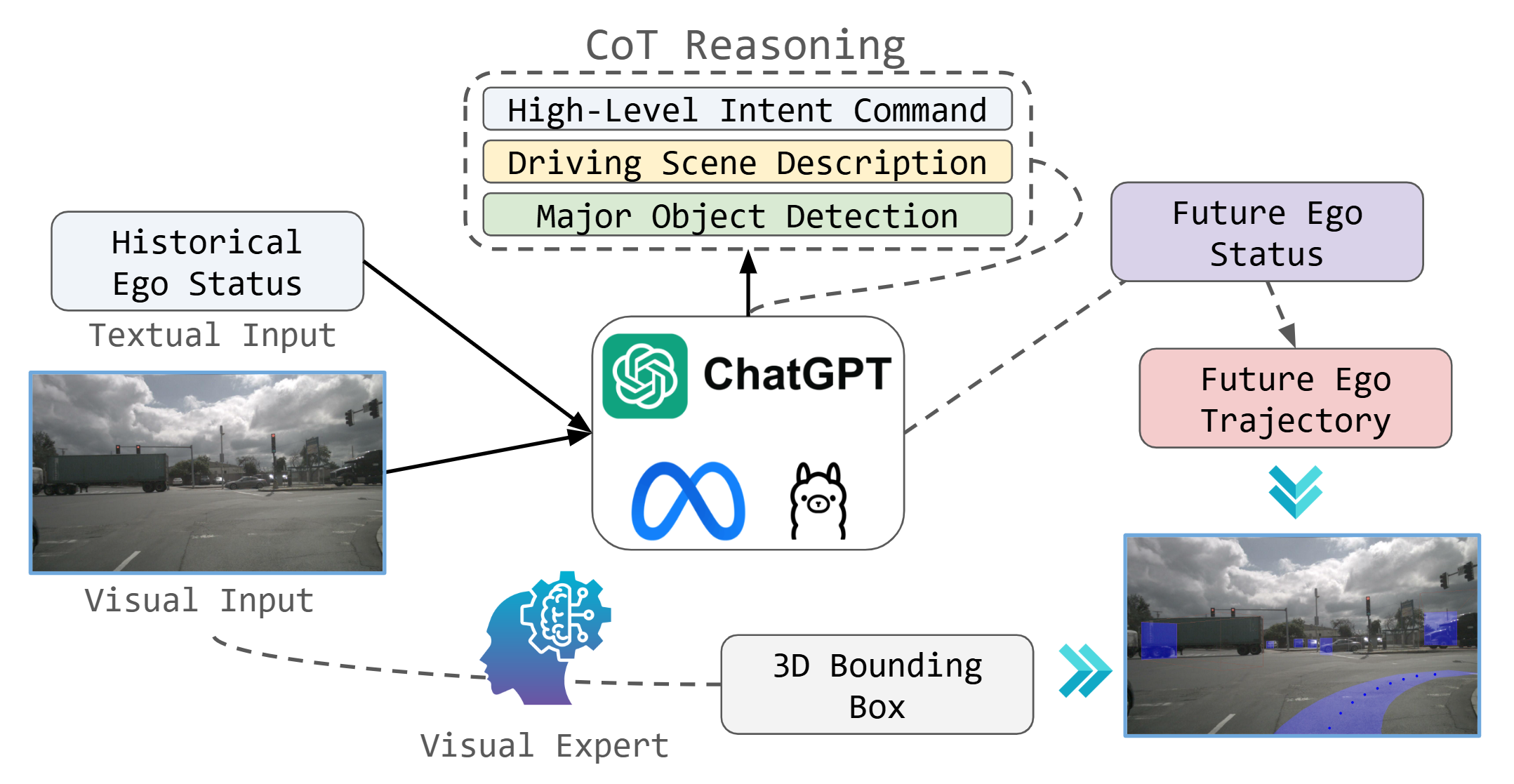}
    \caption{Illustration of the OpenEMMA framework.}
    \label{fig:openemma-pipeline}   
\end{figure*}

To address the limitations of closed-source models like EMMA, we introduce OpenEMMA, an open-source end-to-end AD framework designed to replicate EMMA's core functionalities using publicly available tools and models. Open-EMMA aims to democratize access to these advancements, providing a platform for broader research and development. Similar to EMMA \citep{hwang2024emma}, OpenEMMA processes front-facing camera images and textual historical ego vehicle status as inputs. Driving tasks are framed as Visual Question Answering (VQA) problems, with Chain-of-Thought reasoning employed to guide the models in generating detailed descriptions of critical objects, behavioral insights, and meta-driving decisions. These decisions are directly inferred by the model itself, providing essential context for waypoint generation. To mitigate the known limitations of MLLMs in object detection tasks, Open-EMMA integrates a fine-tuned version of YOLO specifically optimized for 3D bounding box prediction in AD scenarios, significantly enhancing detection accuracy. Additionally, by leveraging the MLLM's pre-existing world knowledge, OpenEMMA can produce interpretable, human-readable outputs for perception tasks such as scene understanding, thereby improving transparency and usability. The complete pipeline and supported tasks are illustrated in Figure \ref{fig:openemma-pipeline}.


We summarize our main contributions as follows:
\begin{itemize}
    \item We introduce \textbf{OpenEMMA}, an open-source end-to-end Multimodal Model for autonomous driving that leverages existing open-source modules and pre-trained MLLMs to replicate the functionalities of EMMA in trajectory planning and perception.
    \item We then perform extensive experiments on the validation set of the nuScenes dataset~\citep{nuscenes2019}, assessing the performance of OpenEMMA with a diverse selection of MLLMs in end-to-end trajectory planning, showcasing its effectiveness and adaptability.
    \item Finally, we fully release the codebase, datasets, and model weights utilized in OpenEMMA in \url{https://github.com/taco-group/OpenEMMA} for the research community to leverage, refine, and extend the framework, propelling further advancements in autonomous driving technology.
\end{itemize}
\section{Methodology}

We develop OpenEMMA, a computing-efficiently End-to-End AD system, based on the pre-trained MLLMs $\mathcal{L}$, as presented in Figure \ref{fig:openemma-pipeline}, predicting the future trajectory $P$ with the historical driving status $T$ and visual driving scenes $I$ as the input as well as detecting the traffic participants.

\subsection{End-to-End Planning with Chain-of-Thought}

Leveraging the power of pre-trained MLLMs~\citep{llama3.2, gpt4o, llavanext, Qwen2VL}, we integrate the Chain-of-Thought reasoning process into the end-to-end trajectory planning process, following an instruction-based approach similar to that in \cite{hwang2024emma}. Because MLLMs are trained to learn human-interpretable knowledge, we prompt them to produce outputs that remain interpretable. Consequently, unlike previous prediction methods that directly generate the trajectory in local coordinates \cite{nayakanti2022wayformermotionforecastingsimple, wang2023eqdrive, multipath}, we instead generate two intermediate representations: the speed vector, $\mathbf{S} = \{s_t\}$, which denotes the magnitude of the vehicle's velocity, and the curvature vector, $\mathbf{K} = \{k_t\}$, representing the turning rate of the vehicle. These presentations aim to reflect how a human drives: speed is how much the gas pedal should be pressed, whereas curvature is how much to turn the steering wheel. 

Given the speed and curvature vectors, we first integrate the heading angle $\theta_t$ at each time step from the product of curvature and speed:

\begin{align*}
       \theta_t &= \theta_{t-1} + \int_{t-1}^{t} k(\tau) s(\tau) \, d\tau,
\end{align*}

We can then compute the velocity components in the $x$ and $y$ directions as: 

\begin{align*}
   v_x(t) &= s_t \cos(\theta_t), \\
   v_y(t) &= s_t \sin(\theta_t).
\end{align*}

Thus, the final trajectory in ego coordinates is computed by integrating the velocity components:
\begin{align*}
       x_t &= x_{t-1} + \int_{t-1}^{t} v_x(\tau) \, d\tau, \\
       y_t &= y_{t-1} + \int_{t-1}^{t} v_y(\tau) \, d\tau,
\end{align*}
with the initial position $(x_0, y_0)$ provided as input. Additionally, for numerical integration, the following cumulative trapezoidal rule is applied:
\begin{align*}
       \theta_t &\approx \theta_0 + \sum_{i=1}^t k_i s_i \Delta t, \\
       x_t &\approx x_0 + \sum_{i=1}^t v_x(i) \Delta t, \\
       y_t &\approx y_0 + \sum_{i=1}^t v_y(i) \Delta t,
\end{align*}
where $\Delta t$ is the time step.

This approach provides a robust and interpretable pathway for planning by decomposing the trajectory generation task into human-interpretable components, mirroring the driving process.

\paragraph{Stage 1: Reasoning:} Initially, we use the front camera image of the driving scene and the historical data (speed and curvature over the past 5 seconds) of the ego car as inputs to the pre-trained MLLMs. Subsequently, we design task-specific prompts to guide the MLLMs in generating comprehensive reasoning of the current ego-driving scenario, covering the following aspects:
\begin{itemize}
    \item \textbf{Intent Command:} A clear articulation of the ego vehicle's intended action based on the current scene, such as whether it will continue following the lane to turn left, turn right, or proceed straight. Additionally, it specifies whether the vehicle should maintain its current speed, slow down, or accelerate.
    \item \textbf{Scene Description}: A concise description of the driving scene according to traffic lights, movements of other cars or pedestrians, and lane markings.
    \item \textbf{Major Objects:} Identify the road users that the ego driver should pay attention to, specifying their location within the driving scene image. For each road user, provide a brief description of their current actions and explain why their presence is important for the ego vehicle's decision-making process. 
\end{itemize}

\paragraph{Stage 2: Predicting} By incorporating the Chain-of-Thought reasoning process and the historical ego status, the MLLMs are prompted to generate the speed $\mathbf{S} = \{s_t\}_{t=0}^T$ and curvature $\mathbf{C} = \{c_t\}_{t=0}^T$ for the next $T$ seconds ($2T$ trajectory points). These predictions are then integrated to compute the final trajectory $\mathbf{T} = \{x_t, y_t\}_{t=0}^T$.

\subsection{Object Detection Enhanced by Visual Sepcialist}
One of the critical tasks in AD is detecting 3D bounding boxes for on-road objects. We observed that off-the-shelf pre-trained MLLMs struggle to deliver high-quality detections due to limitations in spatial reasoning. To overcome this challenge and achieve high detection accuracy without additional fine-tuning of the MLLM, we integrated an external, visually specialized model into OpenEMMA, effectively addressing the detection task.

Our proposed OpenEMMA focuses exclusively on object detection using a front-facing camera and processes data from a single frame, rather than a sequence of consecutive frames. This places the task within the scope of monocular camera-based 3D object detection. Research in this field generally falls into two categories: depth-assisted methods \citep{epropnp, pseudolidar, monodetr} and image-only methods \citep{yolo3d, pix2seq, monodiff, monoedge}. Depth-assisted methods predict depth information to aid detections, while image-only methods rely solely on RGB data for direct predictions. Among these approaches, we selected YOLO3D \citep{yolo3d} for its combination of reliable accuracy, high-quality open-source implementation, and lightweight architecture, which enables efficient fine-tuning and practical integration.

\begin{table*}[htbp]
  \begin{center}
    \begin{tabular}{llccccccccc}
      \toprule
        Method
        & Model
        & L2 (m) 1s
        & L2 (m) 2s
        & L2 (m) 3s
        & L2 (m) avg
        & Failure rate (\%)
        \\

        \midrule

        \multirow{3}{*}{Zero-shot} 
        & LLaVA-1.6-Mistral-7B
        & 1.66
        & 3.54
        & 4.54
        & 3.24
        & 4.06
        \\
        & Llama-3.2-11B-Vision-Instruct
        & 1.50
        & 3.44
        & 4.04
        & 3.00
        & 23.92
        \\
        & Qwen2-VL-7B-Instruct
        & 1.22
        & 2.94
        & 3.21
        & 2.46
        & 24.00
        \\

        \midrule

        \multirow{3}{*}{OpenEMMA} 
        & LLaVA-1.6-Mistral-7B
        & 1.49
        & 3.38
        & 4.09
        & 2.98
        & 6.12
        \\
        & Llama-3.2-11B-Vision-Instruct
        & 1.54
        & 3.31
        & 3.91
        & 2.92
        & 22.00
        \\
        & Qwen2-VL-7B-Instruct
        & 1.45
        & 3.21
        & 3.76
        & 2.81
        & 16.11
        \\ 
      \bottomrule
    \end{tabular}
  \end{center}
  \caption{End-to-end trajectory planning experiments on nuScenes.}
  \label{tab:ade}
\end{table*}

YOLO3D is a two-stage 3D object detection method that enforces a 2D-3D bounding box consistency constraint. Specifically, it assumes that each 3D bounding box is tightly enclosed within its corresponding 2D bounding box. The method begins by predicting 2D bounding boxes and subsequently estimates the 3D dimensions and local orientation of each detected object. The seven parameters of a 3D bounding box---center positions $t_x, t_y, t_z$, dimensions $d_x, d_y, d_z$, and the yaw angle $\theta$---are jointly calculated based on the 2D bounding box and the 3D estimations.

\section{Experiments}\label{sec:exp}

In this section, we first present experiments conducted for end-to-end trajectory planning, utilizing a diverse range of MLLMs to showcase the effectiveness of OpenEMMA. Additionally, we provide detailed insights into the implementation and adaptation of YOLO11n for AD scenarios, emphasizing its seamless integration within the OpenEMMA framework. Finally, we present visual results that highlight OpenEMMA's capabilities in addressing challenging AD scenarios, demonstrating its robustness and effectiveness under diverse conditions.

\subsection{End-to-End Trajectory Planning}\label{sec:exp-e2e}
\paragraph{Setup} The experiments conducted on the validation set of the nuScenes dataset~\citep{nuscenes2019}, and the models tested include GPT-4o~\citep{gpt4o}, LLaVA-1.6-Mistral-7B~\citep{llavanext}, Llama-3.2-11B-Vision-Instruct~\citep{llama3.2}, and Qwen2-VL-7B-Instruct~\citep{Qwen2VL}. For comparison, we use the zero-shot method as the baseline, which relies solely on the historical ego status and the driving scene image, without incorporating any reasoning process. Furthermore, we set $T=5$, prompting the MLLM to predict the future trajectory over the next 5 seconds. Due to budget constraints and the need for reproducibility, the GPT-4o results are only conducted on a limited set of scenes and will be discussed in the case study.
\begin{figure}{}
     \centering
     \begin{subfigure}[b]{0.45\textwidth}
         \centering
         \includegraphics[width=\textwidth]{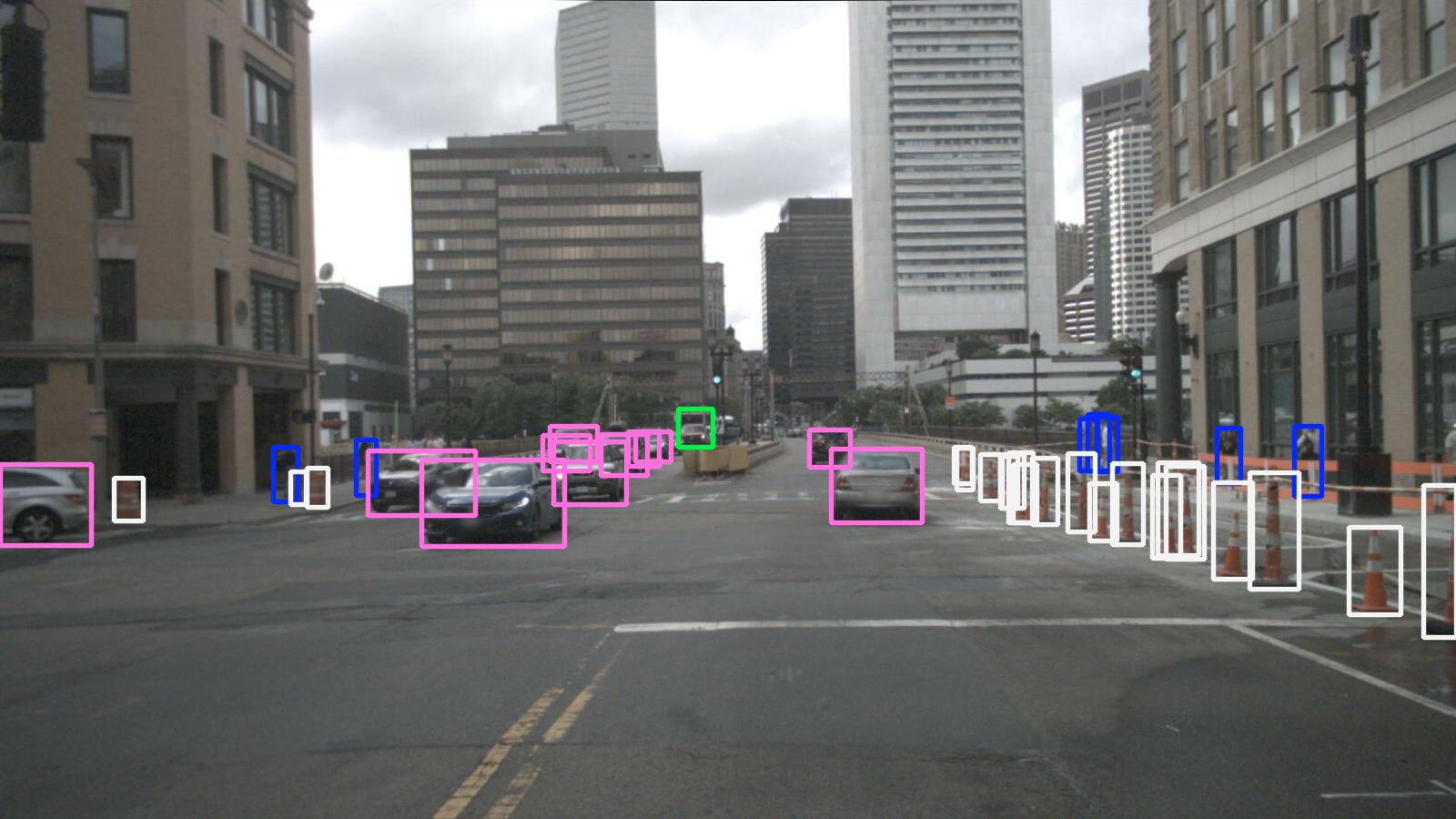}
         \caption{}
     \end{subfigure}
     \hfill
     \begin{subfigure}[b]{0.45\textwidth}
         \centering
         \includegraphics[width=\textwidth]{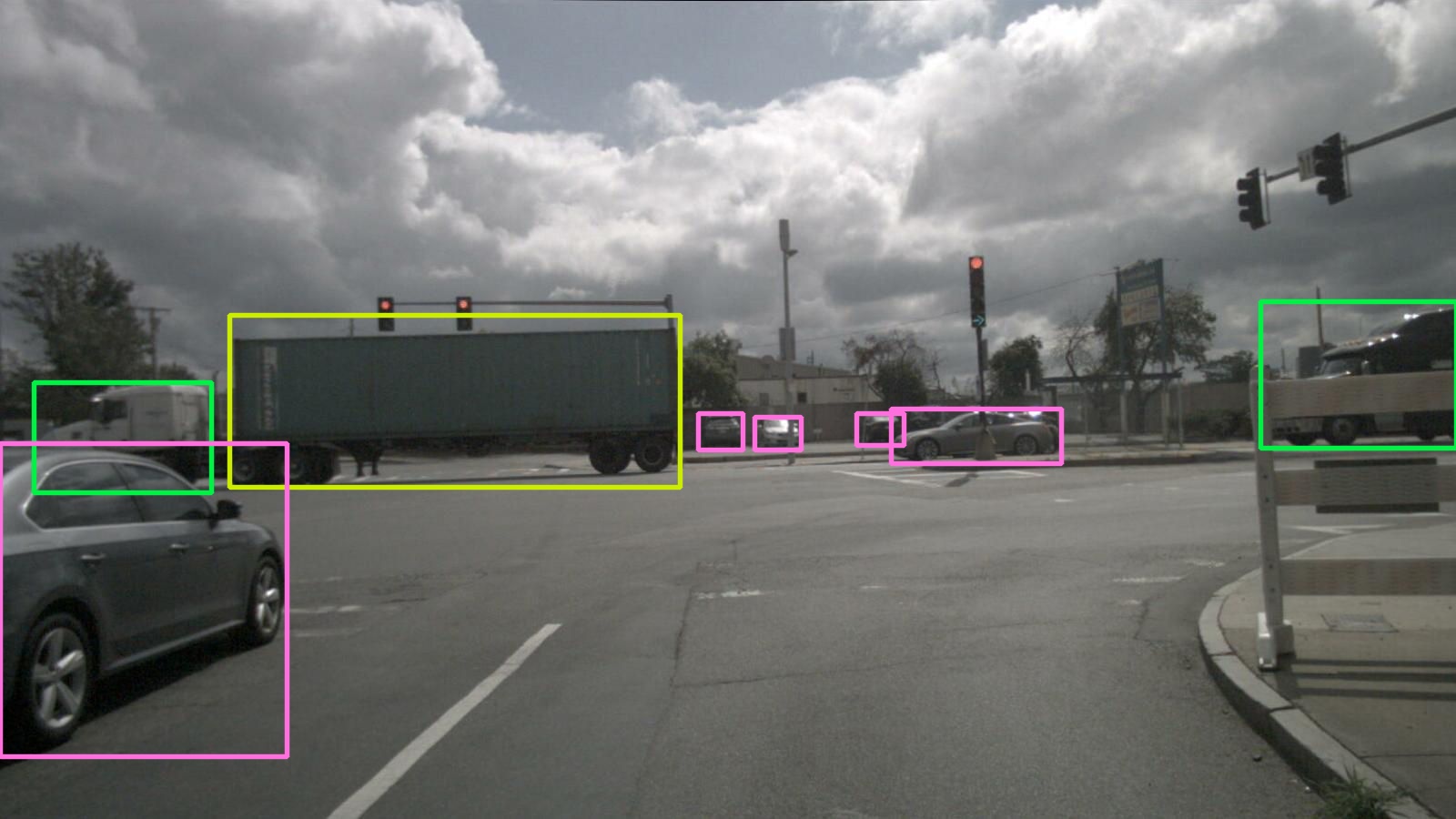}
         \caption{}
     \end{subfigure}
        \caption{YOLO 2D detection results. The class-color correspondences are: cars in pink, trucks in green, trailers in yellow, pedestrians in blue, and traffic cones in white. }
        \label{fig:yolo2d}
\end{figure}
\paragraph{Results} Table \ref{tab:ade} summarizes the performance of OpenEMMA across 150 scenes from the validation set of the nuScenes dataset~\citep{nuscenes2019} in terms of the L2 norm error relative to the ground truth trajectory. Furthermore, a prediction is considered a failure if the L2 norm exceeds 10 within the first second of the future trajectory, and the failure rate is also included in the table. Our key findings are as follows: The overall performance of the MLLMs without fine-tuning for end-to-end trajectory planning is inferior compared to fine-tuning-based approaches, such as those presented in \cite{hwang2024emma}. This outcome is expected, as fine-tuning enables models to better adapt to the specific demands and intricacies of trajectory planning tasks. OpenEMMA consistently outperforms the zero-shot baseline in both L2 norm error and failure rate, demonstrating the effectiveness of the Chain-of-Thought reasoning process in understanding and analyzing complex real-world driving scenarios. Notably, OpenEMMA shows a significant improvement over the zero-shot baseline when using LLaVA-1.6-Mistral-7B as the backbone and a modest yet noticeable enhancement with Llama3.2-11B-Vision-Instruct as the backbone in both L2 norm and failure rate. However, the L2 norm error of OpenEMMA when using Qwen2-VL-7B-Instruct is higher than that of the zero-shot baseline. This is because OpenEMMA successfully generates predictions for many cases where the zero-shot baseline fails. Despite this improvement, it still struggles to produce high-quality trajectories in these challenging scenarios, leading to an overall increase in the L2 norm error. Nevertheless, the significant reduction in failure rate highlights OpenEMMA's improved robustness and capability in handling difficult situations.

\begin{figure*}[htbp]
    \centering
    \begin{subfigure}{\textwidth}
    \includegraphics[scale=0.1]{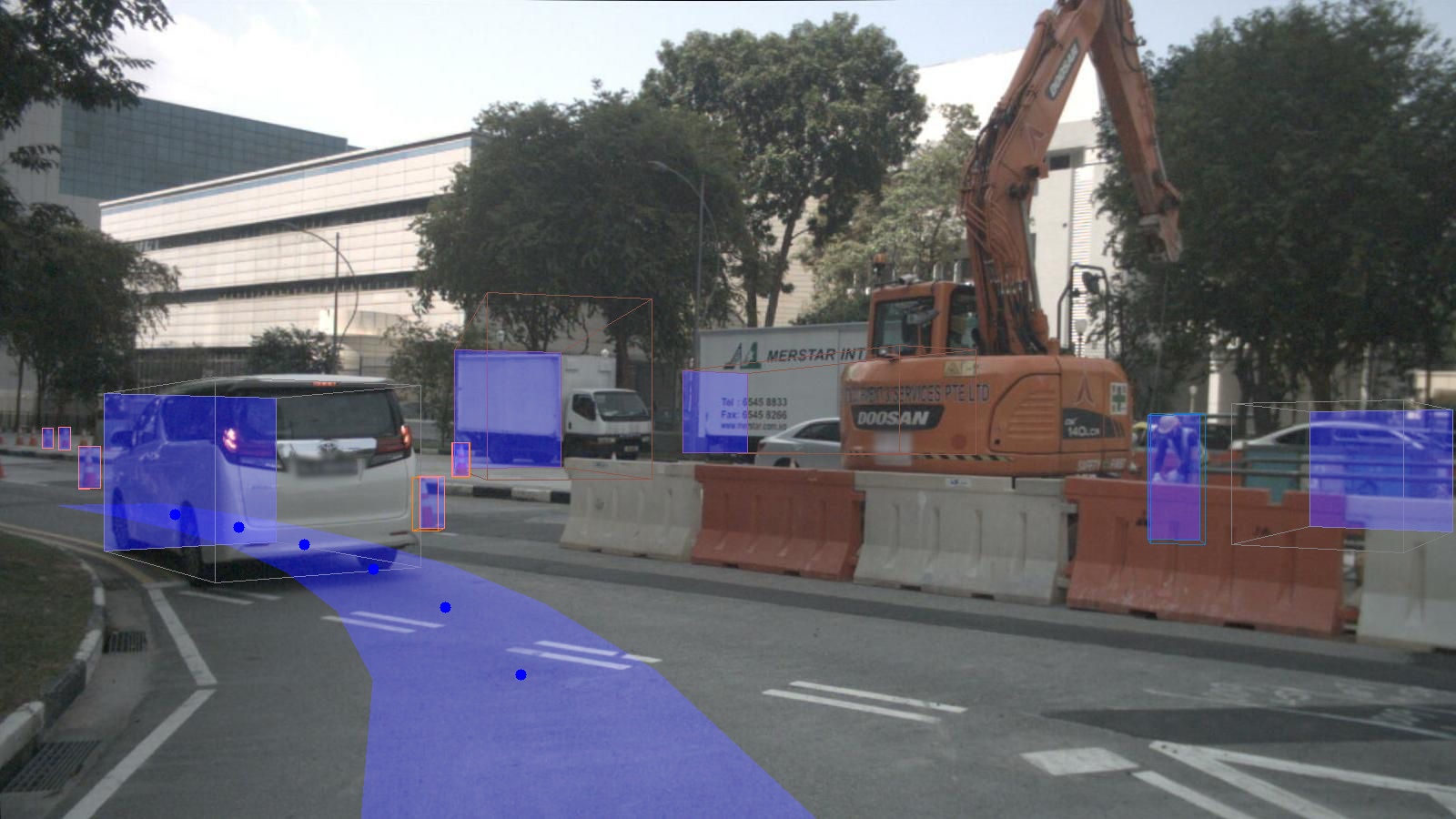}
    \includegraphics[scale=0.1]{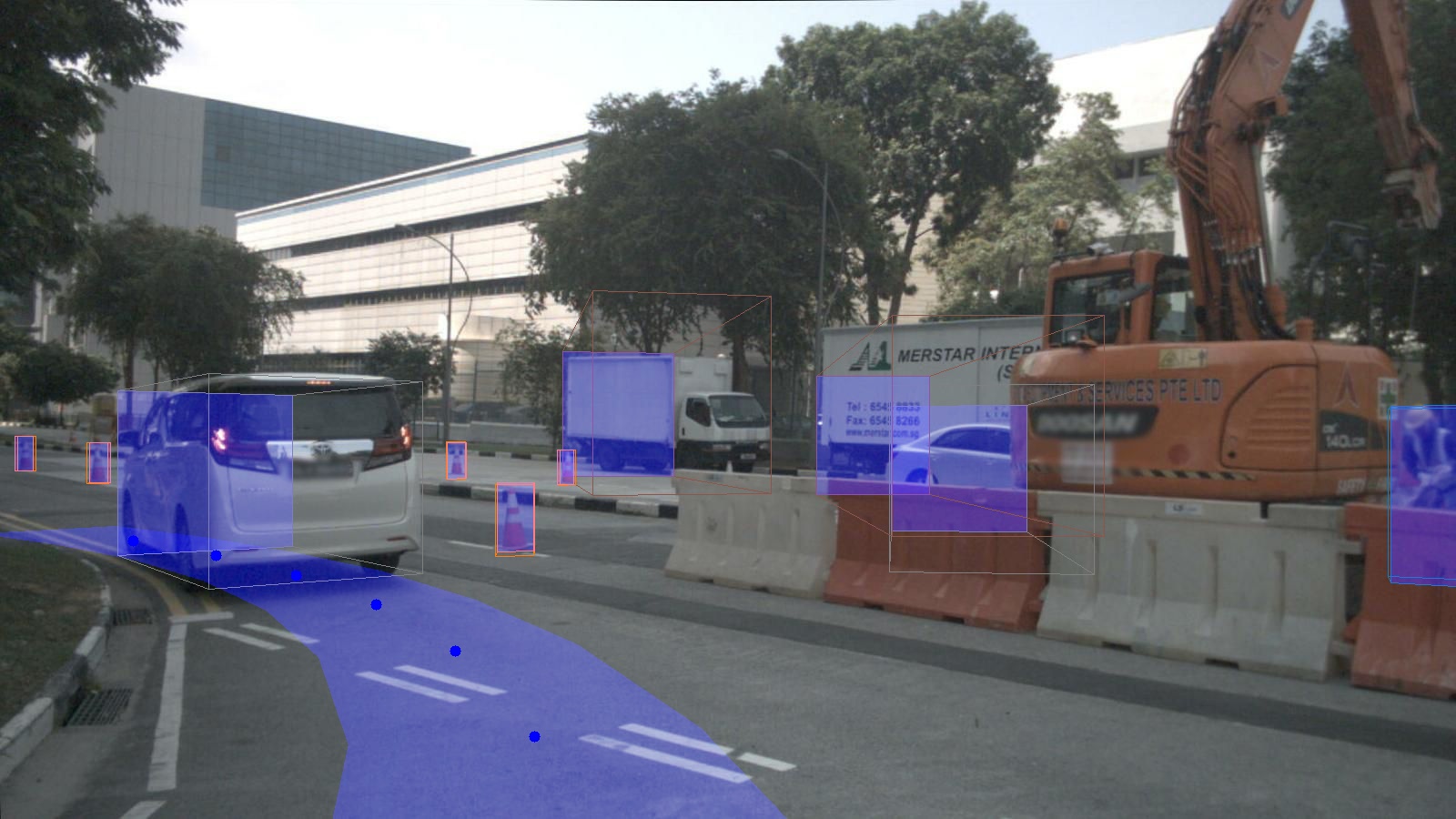}
    \includegraphics[scale=0.1]{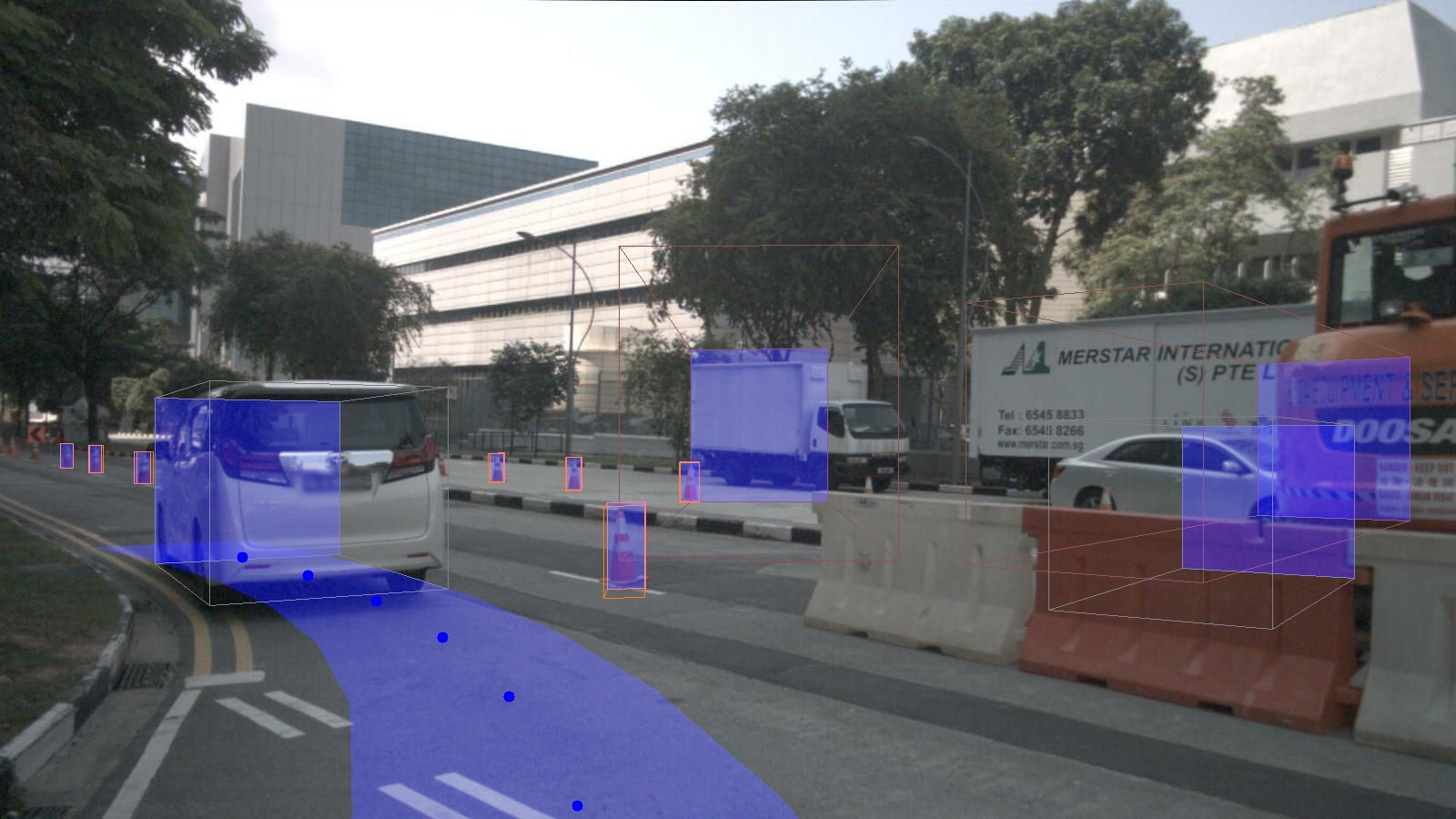}
    \subcaption{Illustration of OpenEMMA's prediction in a right-turn scenario on the road.}
    \label{fig:vis-a}   
    \end{subfigure}

    \begin{subfigure}{\textwidth}
    \includegraphics[scale=0.1]{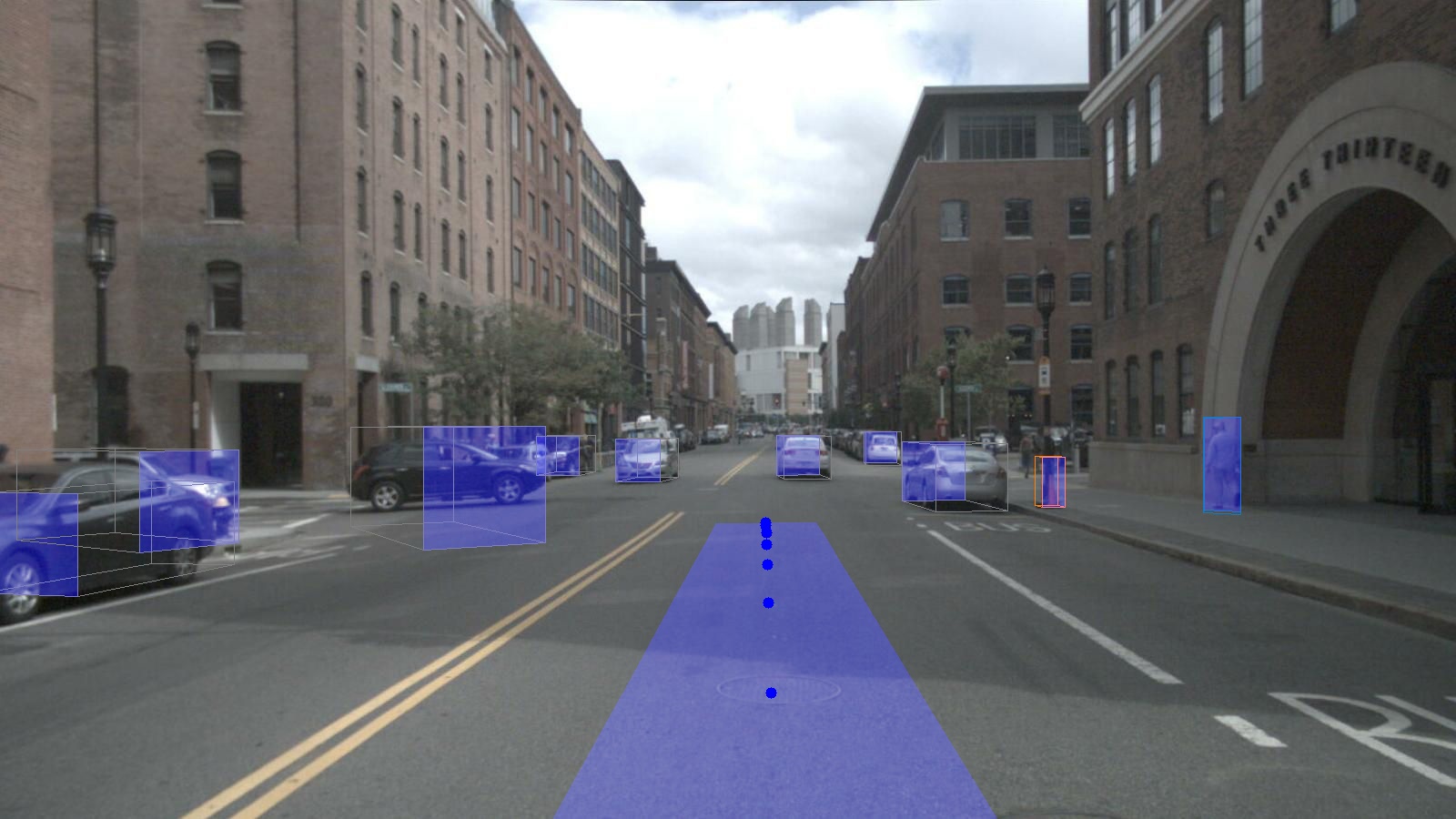}
    \includegraphics[scale=0.1]{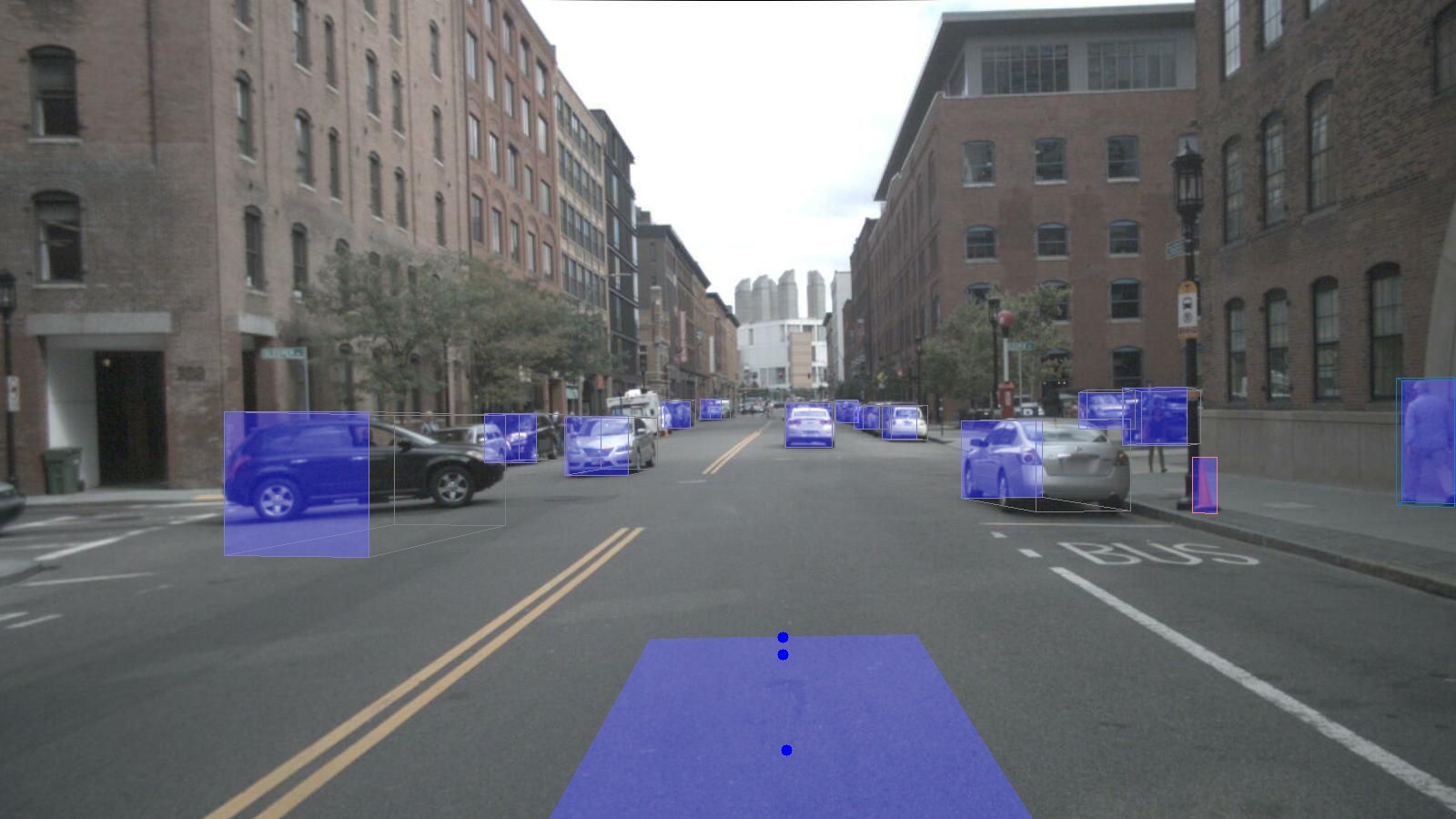}
    \includegraphics[scale=0.1]{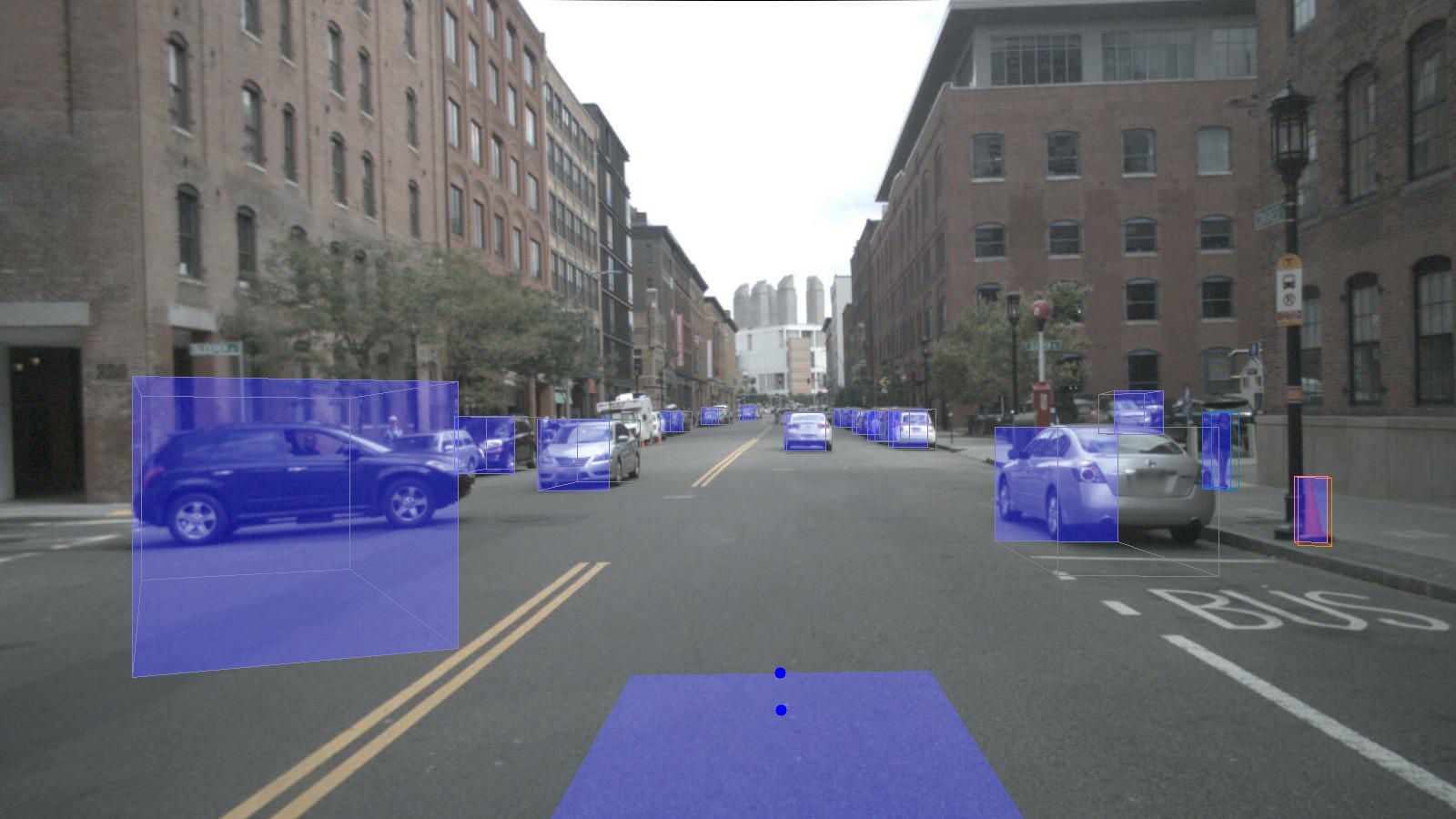}
    \subcaption{Illustration of OpenEMMA's prediction when encountering a left-turning vehicle.}
    \label{fig:vis-b}
    \end{subfigure}

    \begin{subfigure}{\textwidth}
    \includegraphics[scale=0.1]{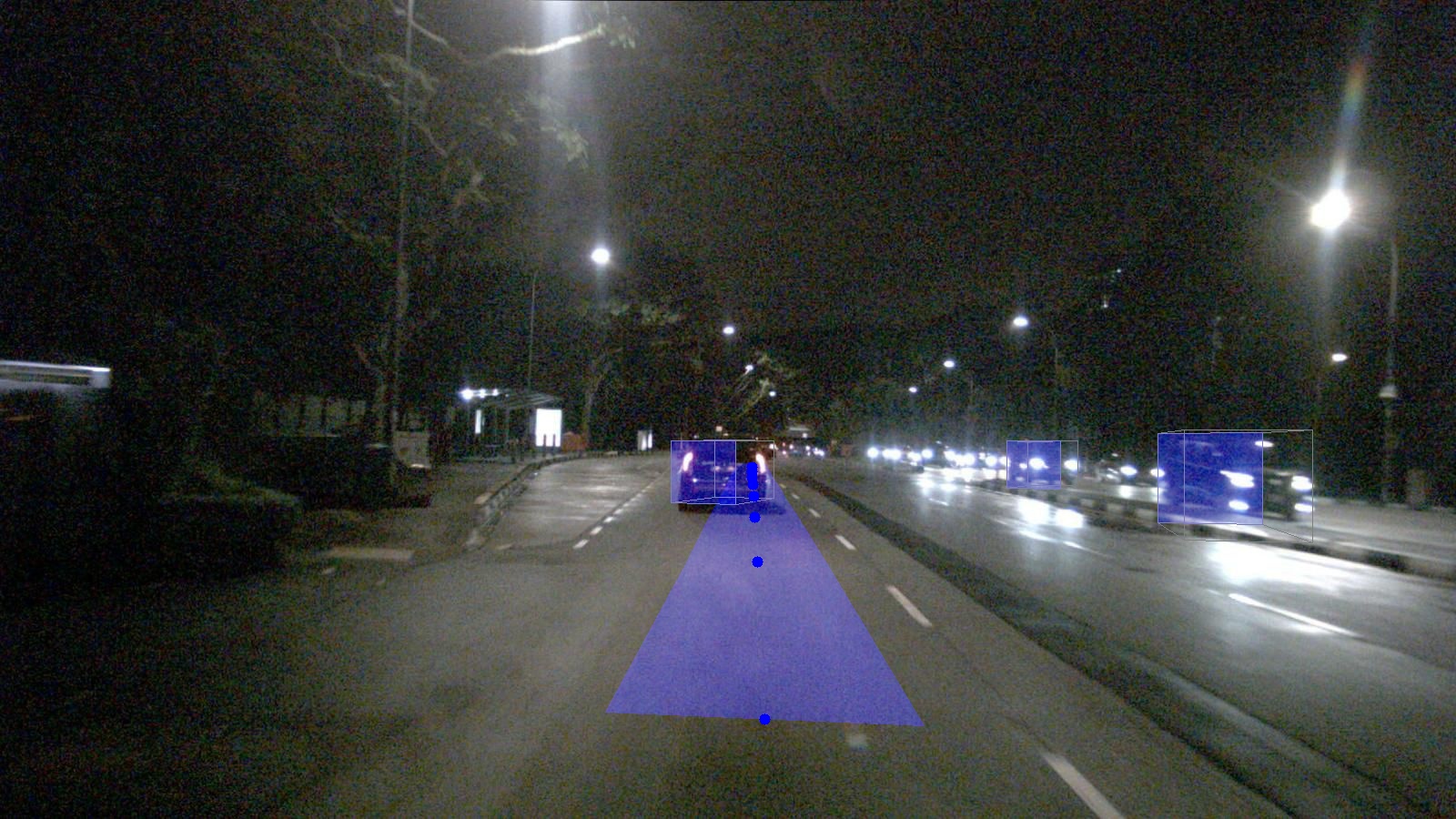}
    \includegraphics[scale=0.1]{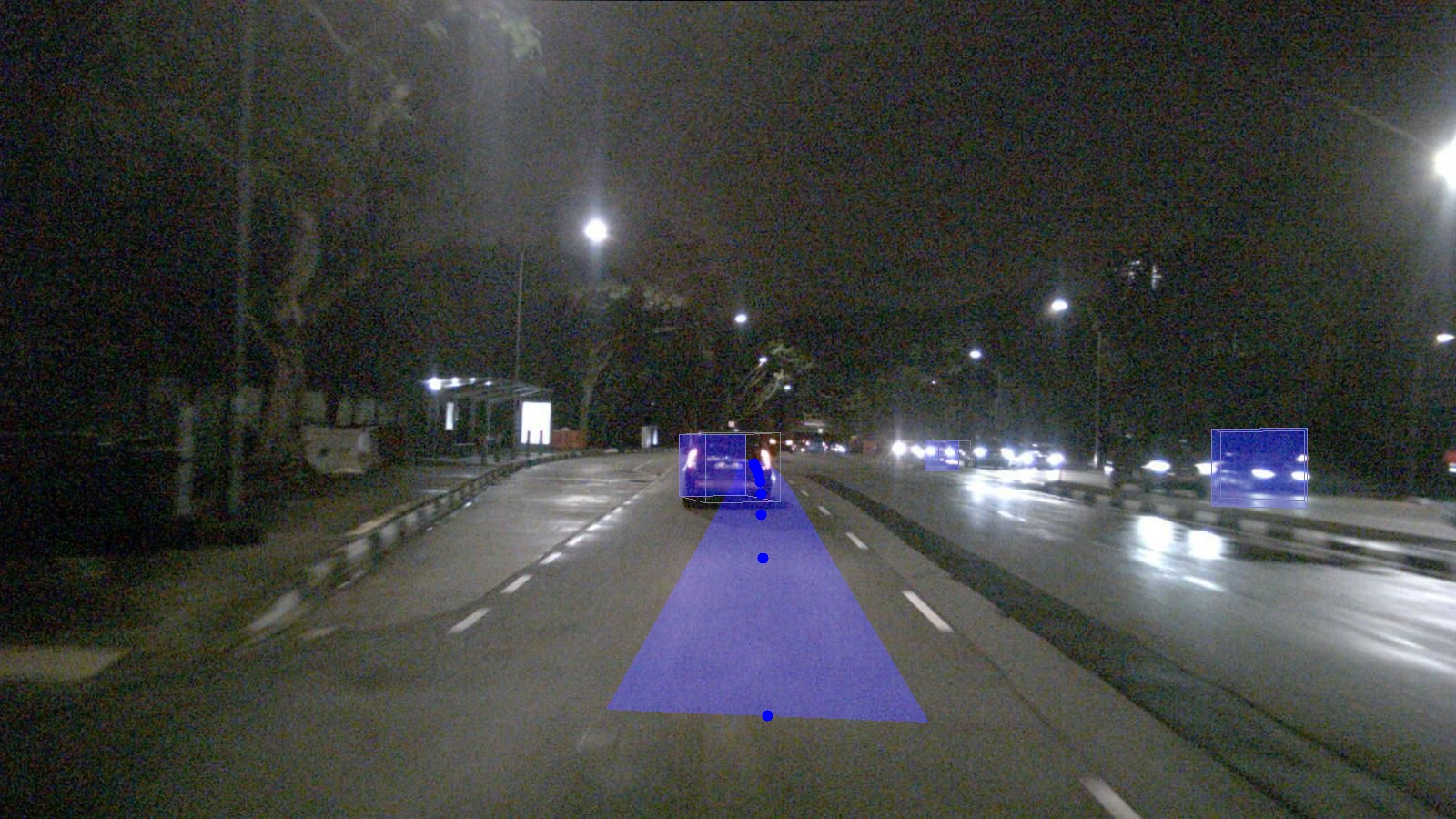}
    \includegraphics[scale=0.1]{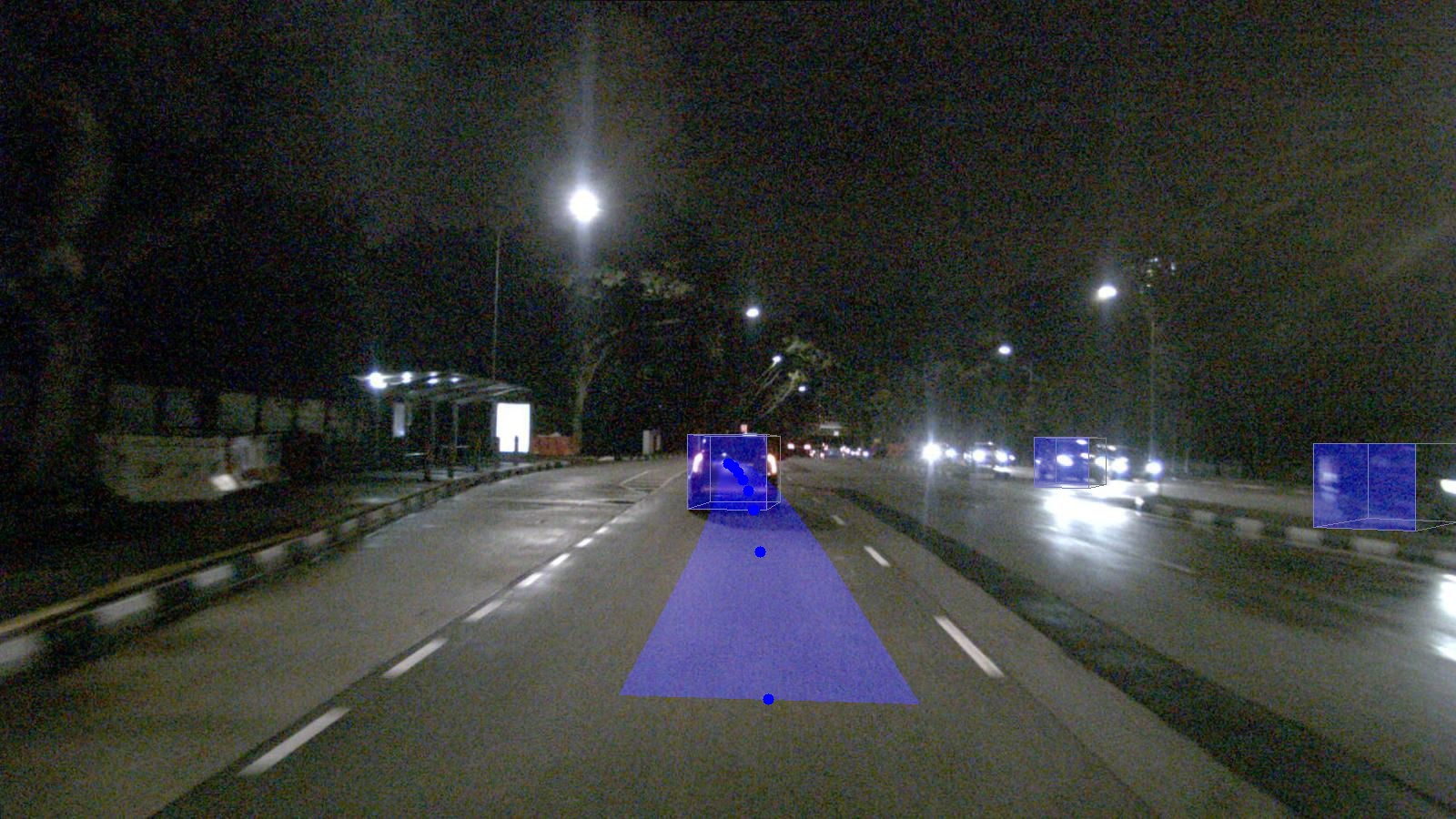}
    \subcaption{Illustration of OpenEMMA's predictions in low-light nighttime conditions.}
    \label{fig:vis-c}
    \end{subfigure}

\caption{Visualization of OpenEMMA predictions powered by GPT-4o.}\label{fig:visual}
    
\end{figure*}
\subsection{3D Object Detection}
Our implementation builds upon the open-source repository\citep{yolo3dgit}, with modifications to replace the 2D detection network with YOLO11n\citep{ultralytics}. The YOLO11n was fine-tuned on the nuImages dataset\citep{nuscenes2019} with images downsampled to $640\times360$. We loaded weights pre-trained on COCO dataset provided by ultralytics\citep{ultralytics}, and trained the network on a single RTX 4060Ti for 300 epochs. The batch size was chosen as 50 and an SGD optimizer was used with a learning rate of 0.01, a momentum of 0.937, and a weight decay of 0.0005. The learning rate decreased linearly to 0.0001 at the end of the training. The best result is achieved at epoch 290 with the mAP50 equal to 0.60316. The weight of the 3D estimation network remains unchanged, utilizing weights from the Yolo3D repository. Figure \ref{fig:yolo2d} illustrates the 2D bounding box detection results from the fine-tuned YOLO11n network. The 3D bounding box detection results are included in the main demonstration videos. 


\subsection{Visualization}
Figure \ref{fig:visual} presents three visual examples from a variety of challenging driving scenarios, highlighting the robustness and effectiveness of OpenEMMA under diverse conditions. In these scenarios, GPT-4o is utilized as the backbone, processing not only the current driving scene but also visual inputs from the past 5 seconds (10 frames). All other settings remain consistent with those described in Section \ref{sec:exp-e2e}.

Figure \ref{fig:vis-a} showcases OpenEMMA's performance during a scenario where the ego vehicle is making a right turn while following the designated lane. OpenEMMA demonstrates its capability to accurately detect on-road objects, plan a smooth and precise trajectory, and adhere to driving rules, ensuring safe and efficient navigation through the turn.

Figure \ref{fig:vis-b} illustrates the visualization of OpenEMMA in a potentially unsafe driving scenario, where a vehicle suddenly enters the current lane from a sharp turn. OpenEMMA promptly detects the risk factor and makes the appropriate decision—to brake and maintain a safe distance, effectively preventing a potential collision. This example highlights OpenEMMA's capability to handle complex driving situations, showcasing its robust reasoning and ability to ensure safety in dynamic and unpredictable environments.

Figure \ref{fig:vis-c} illustrates the performance of OpenEMMA under low-light nighttime conditions. While OpenEMMA may occasionally miss detecting certain objects in such challenging environments, it successfully identifies and detects key objects critical for safe navigation. Moreover, it accurately understands that the ego vehicle is transitioning to the left lane and generates precise trajectory planning to accommodate the maneuver effectively. This demonstrates OpenEMMA's robustness in handling complex driving scenarios with reduced visibility.

\section{Related Work}

\paragraph{End-to-End AD} A significant trend in autonomous driving is the emergence of end-to-end systems~\citep{chen2024end}, which offer increased efficiency by seamlessly transferring feature representations across system components. This contrasts with traditional methods, as the entire system is optimized for the driving task, leading to improved computational efficiency and consistency through shared backbones. These end-to-end approaches can be broadly divided into imitation learning and reinforcement learning. Within reinforcement learning, models like Latent DRL~\cite{toromanoff2020end}, Roach~\cite{zhang2021end}, and ASAP-RL~\cite{wang2023efficient} prioritize enhancing decision-making. Complementarily, models like ScenarioNet~\cite{li2024scenarionet} and TrafficGen~\cite{feng2023trafficgen} focus on generating diverse driving scenarios to improve system robustness during testing. More recently, MLLMs have been integrated into autonomous driving systems. For example, LMDrive~\cite{shao2024lmdrive} facilitates natural language interaction and advanced reasoning, enabling more intuitive human-vehicle communication. Senna~\cite{jiang2024sennabridginglargevisionlanguage} takes this further by combining MLLMs with end-to-end systems, decoupling high-level planning from low-level trajectory prediction. Building upon these developments, EMMA~\citep{hwang2024emma}, powered by Gemini, represents a significant step forward. This vision-language model transforms raw camera sensor data into diverse driving-specific outputs, including planner trajectories, perceived objects, and road graph elements, showcasing the potential of MLLM integration for enhanced functionality and efficiency in autonomous driving.

\paragraph{MLLM for AD} Multimodal Large Language Models (MLLMs)~\citep{li2022blip, li2023blip2, liu2024llava,llavanext,llama3.2, Qwen-VL, Qwen2VL} extend the capabilities of Large Language Models (LLMs)~\citep{devlin2018bert, radford2019gpt2,brown2020gpt3,team2023gemini,roziere2023codellama,touvron2023llama,touvron2023llama2, raffel2020t5,qwen2,qwen2.5} into the visual realm. LLMs, known for their generalizability, reasoning, and contextual understanding, provide the foundation upon which MLLMs are built. The key to enabling MLLMs to seamlessly process both textual and visual information lies in aligning visual and text embeddings. This is achieved by using vision encoders, such as CLIP~\citep{radford2021clip}, to convert image patches into visual tokens that are aligned with the text token space, thereby unlocking new possibilities for comprehensive multimodal understanding.

MLLMs have been widely applied in real-world scenarios, particularly in the field of autonomous driving. GPT-Driver~\citep{mao2023gptdriverlearningdrivegpt} transforms both the planner inputs and outputs into language tokens. By utilizing GPT-3.5, it generates driving trajectories described through natural language representations of coordinate positions. DriveVLM \citep{tian2024drivevlm} utilizes Chain-of-Thought (CoT) \citep{wei2023chainofthoughtpromptingelicitsreasoning} for advanced spatial reasoning and real-time trajectory planning. RAG-Driver~\citep{yuan2024rag} introduces a novel in-context learning approach to AD based on retrieval-augmented generation with MLLMs, enhancing generalizability and explainability in AD systems. Driving-with-LLMs~\citep{chen2024driving} introduces a novel paradigm for fusing the object-level vectorized numeric modality into LLMs with a two-stage pretraining and finetuning method. DriveLM \citep{sima2023drivelm} developed an end-to-end MLLM in AD by leveraging graph-structured Visual Question Answering (VQA) for tasks across perception, prediction, and planning.

\section{Conclusion}
In this paper, we propose OpenEMMA, an open-source, computationally efficient end-to-end autonomous driving framework built on Multimodal Large Language Models. Leveraging historical ego-vehicle data and images captured by the front camera, OpenEMMA employs a Chain-of-Thought reasoning process to predict the future speed and curvature of the ego vehicle, which are then integrated into the trajectory planning process. Additionally, by incorporating a fine-tuned external visual specialist model, OpenEMMA achieves precise detection of 3D on-road objects. Furthermore, the proposed OpenEMMA framework demonstrates significant improvements over zero-shot baselines, showcasing its effectiveness, generalizability, and robustness across various challenging driving scenarios.

\section{Limitation, and Future Work} 
As an initial step in developing an end-to-end autonomous driving framework based on off-the-shelf pre-trained models, we incorporated only basic Chain-of-Thought reasoning during inference. While this serves as a foundational approach, there is significant untapped potential to enhance the framework by integrating more advanced inference-time reasoning techniques, such as CoT-SC~\citep{wang2022self} and ToT~\citep{yao2023tot}, into the framework, which could yield more practically effective methods for autonomous driving. 

Furthermore, due to the limited object grounding capabilities of current MLLMs, we incorporated a fine-tuned YOLO model into OpenEMMA to handle object detection tasks, rather than relying solely on the capabilities of the MLLM itself. While this approach provides a practical solution, it highlights the need for future advancements in MLLMs to bridge the gap in spatial reasoning and grounding accuracy. Addressing these limitations will be essential to achieve a truly unified framework that leverages MLLMs for all key perception and reasoning tasks in autonomous driving.

{\small
\bibliographystyle{ieee_fullname}
\bibliography{main}
}

\end{document}